\documentclass[wcp]{jmlr}

\usepackage{longtable}

\usepackage{booktabs}

\usepackage{algcompatible}

\def\W{\mathbf{W}}

\def\X{\mathbf{X}}

\def\real{\rm I\!R}

\def\U{\mathbf{U}}
\def\u{\mathbf{u}}
\def\V{\mathbf{V}}
\def\v{\mathbf{v}}
\def\SSigma{\boldsymbol{\Sigma}}
\DeclareMathOperator*{\argmax}{argmax} 

\usepackage[textsize=tiny]{todonotes}
\usepackage{makecell}

\makeatletter
\let\Ginclude@graphics\@org@Ginclude@graphics 
\makeatother

\jmlrvolume{}
\jmlryear{2022}
\jmlrworkshop{ACML 2022}

\title[]{Training Acceleration of Low-Rank Decomposed Networks\\ using Sequential Freezing and Rank Quantization}

  \author{\Name{Habib Hajimolahoseini} \Email{habib.hajimolahoseini@huawei.com}\\
  \Name{Walid Ahmed} \\
  \Name{Yang Liu} \\
  \addr Toronto Research Center, Huawei Technologies}

\begin{document}

\maketitle

\begin{abstract}
Low Rank Decomposition (LRD) is a model compression technique applied to the weight tensors of deep learning models in order to reduce the number of trainable parameters and computational complexity.
However, due to high number of new layers added to the architecture after applying LRD, it may not lead to a high training/inference acceleration if the decomposition ranks are not small enough. 
The issue is that using small ranks increases the risk of significant accuracy drop after decomposition. 
In this paper, we propose two techniques for accelerating low rank decomposed models without requiring to use small ranks for decomposition.
These methods include rank optimization and  sequential freezing of decomposed layers. 
We perform experiments on both convolutional and transformer-based models.
Experiments show that these techniques can improve the model throughput up to 60\% during training and 37\% during inference when combined together while preserving the accuracy close to that of the original models.
\end{abstract}

\section{Introduction}
During the past few years, deep learning models have become larger and larger with millions or even billions of trainable parameters.
Training such huge models is a computationally expensive and time consuming process which takes a huge portion of memory as well \citep{dean2012large, hajimolahoseini2023methods}. 
On edge devices and smart phones, memory consumption and computational complexity are even more concerning because they can cause several issues regarding memory and battery life \citep{hajimolahoseini2019deep, hajimolahoseini2018ecg, hajimolahoseini2022long}.
However, due to high redundancy in AI models, they could be compressed to the point that the accuracy is preserved close to the original model \cite{chen2019drop}. 

The computational complexity and memory consumption of deep learning models are dominated by the convolutional and fully connected layers, respectively \cite{cheng2018recent}. 
The existing techniques for training/inference acceleration of AI models can be categorized into four different groups
including Low Rank Decomposition (LRD), pruning, quantization and knowledge distillation (KD) \citep{cheng2017survey}. 
In all of these techniques except KD, the original architecture of the model is preserved and only the training layers are compressed so that the memory usage and/or computational complexity is minimized. 
In KD however, the compressed model may have a totally different architecture \citep{hinton2015distilling, rashid2021mate}. 
``For more information about knowledge distillation, the reader is referred to \citep{hinton2015distilling, rashid2021mate}.

In pruning, the sparsity of the model is increased by removing the parts that are not contributing much to the performance of the network \cite{luo2018thinet}. 
It can be applied in different levels including filters, kernels, vectors or weights of the layers \cite{zhang2018systematic, zhuang2018discrimination, mao2017exploring}. 
The pruning is applied according to a heuristic ranking measure which is introduced manually based on experiments.  
Although it can be used as a compression technique, one big challenge of pruning is that it may take a long time for sequences of pruning and fine-tuning to reach to the desired performance which may cause a significant overhead during training \citep{cheng2017survey}. 

In quantization approach on the other hand, the weights or activations of the model are quantized using scalar or fixed-point quantization techniques \cite{cheng2017quantized}. 
In fixed-point weight quantization, the weights are represented using a lower precision e.g. 16-bits, 8-bits or binary \citep{prato2019fully, bie2019simplified}. 
In scalar quantization, the weights are represented using a codebook of centres and codes for assigning to them \cite{cheng2017survey}. 
Knowledge Distillation (KD) uses a teacher-student framework in order to transfer the knowledge from a larger network (teacher) into a compact and efficient one (student) by adding and auxiliary loss to imitate softmax outputs or logits from the teacher as a representation of class distributions \citep{hinton2015distilling, rashid2021mate}.

A common issue with most of the aforementioned techniques is that these methods are based on some heuristics and therefore they have poor or no mathematical support and thus the a closed form solution does not exists.
Furthermore, they could face some serious limitations in high compression ratios.
In contrast with the aforementioned methods, Low Rank Decomposition (LRD) decomposes the weight tensors using a tensor decomposition algorithm e.g. Singular Value Decomposition (SVD) \cite{hajimolahoseinicompressing, hajimolahoseinistrategies, van1987matrix, li2021short}.
In this approach, each fully connected layer is replaced with two consecutive fully connected layers whose weights are calculated from the original matrix using SVD algorithm. 
On the other hand, for convolutional layers, a higher order version of SVD e.g. Tucker is applied in order to decompose them into multiple components \cite{de2000multilinear, de2000best, walid2022speeding}.

Despite several benefits that LRD provides, including a strong mathematical foundation and one-shot knowledge distillation from original to decomposed model, the high number of new layers generated by applying tensor decomposition prevents it from being considered as a training/inference acceleration method in terms of frame per second.
Therefore, it is mostly considered as a type of model compression technique which helps in terms of memory consumption.
A naive way of improving the training/inference speed is to reduce the decomposition ranks to the point that the acceleration is achieved but this may harm the accuracy of the model to the point that it could not be recovered close to the original model. 

In this paper, we show that appropriate rank selection and sequential freezing of the decomposed layers can help improving the efficiency of the LRD decomposed models without requiring to decrease the rank of decomposition significantly. 
In rank optimization technique, we search for the optimal rank around the calculated decomposition ranks in order to find the most efficient decomposed architecture.
A sequential layer freezing of the decomposed layers is also proposed for saving time during back propagation.  
We also show that the proposed techniques are platform-agnostic and could be used in different AI processors e.g. NVIDIA's GPUs and Huawei's Ascend NPUs to improve the training/inference speed.

\section{Proposed Methods}

Each convolutional layer consists of a tensor of trainable parameters that could be represented as $\W\in \real^{C\times S\times h\times w}$, where $C$ and $S$ represent the number of input and output channels, respectively while $h$ and $w$ are the spatial dimensions of the kernels. 
For $1\times 1$ convolutional and fully connected layers, we have $h=w=1$. 
Thus, $\W$ is reduced to the two dimensional matrix $\W\in \real^{C\times S}$. 

We use Singular Value Decomposition (SVD) to decompose $1\times 1$ convolutional and fully connected layers as follows \cite{hajimolahoseinicompressing}:
\begin{equation}\label{svd}
\W = \U\SSigma \V^\top=\sum_{i=1}^R{\sigma_i\u_i\v_i^\top},
\end{equation}
in which $\U \in \real^{C\times C}$ and $\V \in \real^{S\times S}$ contain the eigenvectors and $\SSigma \in \real^{C\times S}$ is a diagonal rectangular matrix containing the singular values $\sigma_i>0$ of $\W$ and $R=\min(C,S)$ is called the rank of $\W$.
In order to achieve compression, we use a smaller rank $r<R$ which leads to the approximation matrix $\W'$, in which only the first $r$ components are used:
\begin{equation}\label{svd2}
\W' = \sum_{i=1}^r{\sigma_i\u_i\v_i^\top}=\U'\SSigma' \V'^\top
\end{equation}
where $\U' \in \real^{C\times r}$ and $\V' \in \real^{S\times r}$ are the new  orthogonal matrices and $\SSigma' \in \real^{r\times r}$ is the new diagonal rectangular matrix.
This matrix is called the low rank approximation of the original matrix $\W$ and the reconstruction error is defined as:
\begin{equation}\label{reconstruct}
e_r = ||\W-\W'||^2
\end{equation}

\begin{figure}
    \begin{minipage}{.48\textwidth}
        \centering
        \includegraphics[width=60mm,scale=1]{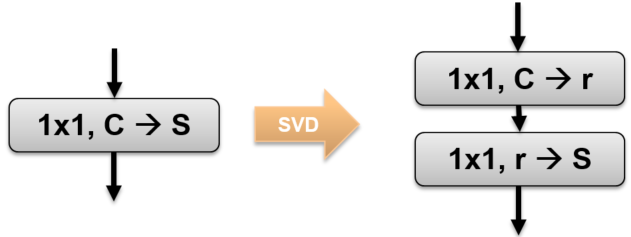}
        \label{fig:1a}
    \end{minipage}
    \begin{minipage}{.48\textwidth}
        \centering
        \includegraphics[width=60mm,scale=1]{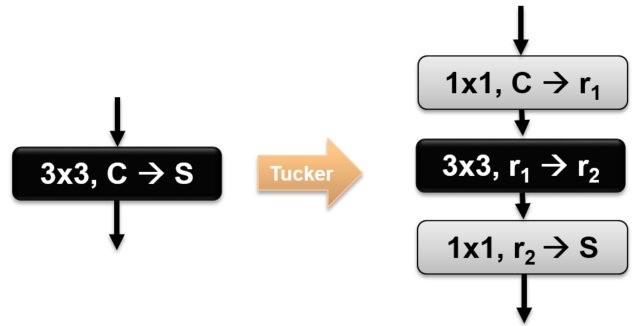}
        \label{fig:1b}
    \end{minipage}
    \caption{Low Rank Decomposition of 1x1 and 3x3 convolutional layers. Note that FC layers are treated the same as 1x1 Conv layers. }
    \label{svd_tucker}
\end{figure}

For regular convolutional layers, a higher order version of SVD e.g. Tucker could be applied in order to decompose them into multiple components.
In case of Tucker, the convolutional layers are decomposed into three layers, a $1\times1$ followed by a $3\times3$ and then another $1\times1$ convolutional layer. 
Assuming that vertical and horizontal kernel sizes are the same $h=w=k$ we represent $\W\in \real^{C\times S \times h \times w}$ as $\W\in \real^{C\times S \times k^2}$.
Therefore, the Tucker decomposition can be applied as follow \cite{gusak2019musco}:
\begin{equation}\label{tucker2}
\W = \X \times_{r_1} \U \times_{r_2} \V
\end{equation}
where $\U \in \real^{C\times r_1}$ and $\V \in \real^{S\times r_2}$ are truncated unitary matrices and $\X \in \real^{r_1\times r_2 \times k^2}$ is the core tensor, containing the truncated 1-mode, 2-mode and 3-mode singular values of $\W$. 
Symbols $\times_{r_1}$ and $\times_{r_2}$ also represent multilinear products between each matrix and the core tensor along dimensions $r_1$ and $r_2$, respectively.
SVD and Tucker decomposition are represented in Fig.~\ref{svd_tucker} when applied to a fully connected, $1\times1$ and regular convolution. 

A naive way of reaching to higher acceleration would be to decrease the rank until the desired speed-up is achieved. 
However, according to \eqref{svd2}, the smaller the rank $r$, the larger the reconstruction error. 
Therefore, we may reach to the point where the accuracy could not be recovered after applying the LRD. 
In the next sections, we propose two techniques that could be used for training/inference acceleration without requiring to reduce the ranks significantly. 

\subsection{Rank Optimization}
It could be shown that because of the low level implementation of the model components e.g. convolutions and fully connected layers on hardware, some specific filter dimensions may be more efficient on AI processing devices such as GPUs. 
That is why all convolutional layers in well known architectures e.g. ResNet have dimensions that are powers of 2 e.g. 256, 512, etc.
However, after the weight tensors are decomposed according to a desirable compression ratio, the estimated ranks may be odd numbers which makes the filter dimensions not optimal in low level calculations.

For example, a convolutional layer with filter dimensions of $[512,512,3,3]$ will be decomposed into 3 convolutional layers of dimensions $[512,309]$, $[309,309,3,3]$ and $[309,512]$, respectively by applying Tucker decomposition with 2x compression. 
Having tensors with dimensions 309 may not lead to efficient calculations on hardware and therefore the training/inference speed-up may not be that significant. 

\begin{figure}
    \begin{minipage}{.48\textwidth}
        \centering
        \includegraphics[width=70mm,scale=1]{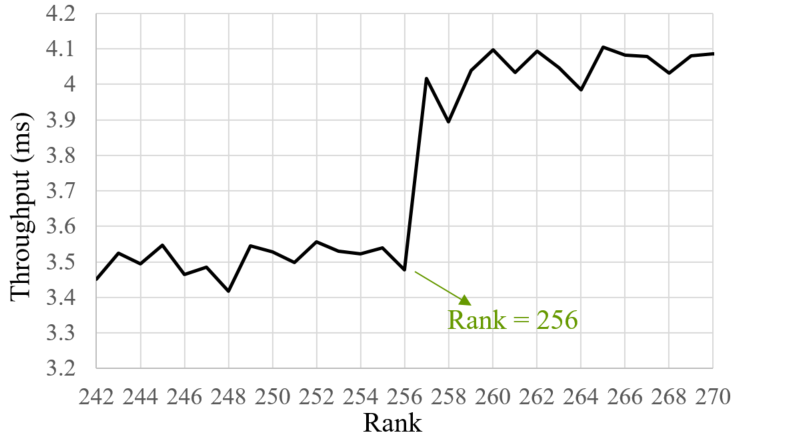}
        \label{fig:2a}
    \end{minipage}
    \begin{minipage}{.48\textwidth}
        \centering
        \includegraphics[width=70mm,scale=1]{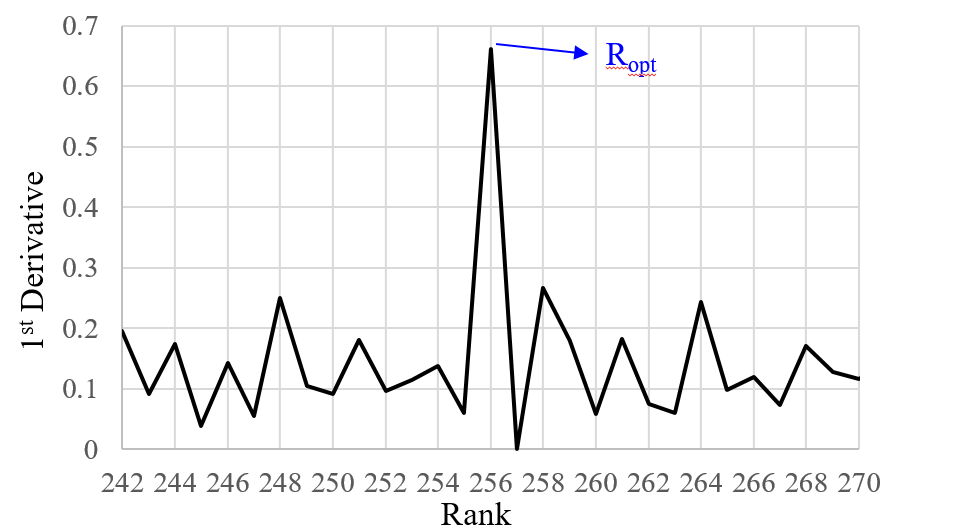}
        \label{fig:2b}
    \end{minipage}
    \caption{Effect of rank selection on throughput of a 3x3 Conv layer in ResNet-152 with dimensions [512, 512, 3, 3] when decomposed using Tucker2 method with different ranks}
    \label{rank_selection}
\end{figure}

Fig.\ref{rank_selection} shows the step time of a convolutional layer versus the corresponding decomposition rank. 
As seen in the top figure, the step time does not change proportional to rank as we decrease the rank from 270 to 242. 
Therefore we may not achieve higher speeds as we decrease the rank or increase the compression ratio. 
However, the layer throughput improves by $15\%$ if we reduce the rank from 257 to 256 while the compression ratio stays almost the same (it changes less than $1\%$). 

Thus, we propose a rank optimization algorithm which finds the optimal ranks around the LRD estimated ranks. 
As shown in Fig.\ref{rank_selection}, these optimal ranks can be spotted by finding the local maximum points the  first derivative curve of the step time vs rank.
In this method, we first generate this curve around the estimated rank by decreasing the rank incrementally until a lower compression ratio is achieved. 
The first peak of the first derivative of the curve is chosen as the optimum rank. 

We also compare the throughput of the decomposed layer with that of the original layer to make sure that using the optimum rank will result in a faster implementation comparing to the original layer. 
If the original layer is still faster, we use the original layer. 
Otherwise, the decomposed layer with the optimum rank is replaced with the original layer. 
This process is repeated for each of the layers independently. 
The proposed algorithm is explained as a pseudo code in Algorithm \ref{rank-selection}. 

\begin{algorithm}
\caption{Find the rank $R_{opt}$ that leads to more efficient computations}
\label{rank-selection}
    \begin{algorithmic} 
    \STATE \textbf{Input}: Original layer $L$, Rank $R$, Lower bound rank $R_{\min}$, Input tensor $x$
    \STATE $T \leftarrow$ Processing time of original layer: $y = L(x)$
    \STATE \textbf{Initialization}: $r \leftarrow R$ and $R_{opt} \leftarrow 0$
    \WHILE{$r \geq R_{\min}$}
        \STATE $L_{r} \leftarrow$ Decompose layer $L$ using rank $r$
        \STATE $t(r) \leftarrow$ Processing time of decomposed layer: $y=L_{r}(x)$
            \IF{$r < R$}
                \STATE $\Delta t(r) \leftarrow t(r)-t(r-1)$
            \ENDIF
        \STATE $r \leftarrow r-1$
    \ENDWHILE
    \STATE \textbf{Optimal Rank:} $R_{opt} \leftarrow \argmax\limits_{r \in [R_{\min}, R]} \Delta t(r)$
    \IF{$t(R_{opt}) < T$}
        \STATE Replace $L$ with $L_r$ 
    \ELSE
        \STATE Use original layer $L$
    \ENDIF
    \end{algorithmic}
\end{algorithm}


In a convolutional layer $\W\in \real^{C\times S\times h\times w}$, assuming that the kernels are prime i.e. $h=w=k$, and $r_2=\beta r_1$, the rank $r_1$ could be calculated as follow to achieve a compression ratio of $\alpha$:
\begin{equation}\label{tucker_rank}
r_1=\frac{-\frac{C+\beta S}{\beta k^2} + \sqrt{\frac{(C+\beta S)^2}{\beta^2k^4} + \frac{4CS}{\beta \alpha}}}{2}
\end{equation}
For rank optimization using Algorithm \ref{rank-selection}, we start from the ranks calculated by \eqref{tucker_rank} for the desired compression ratio of $\alpha$.
Then we calculate $R_{min}$ using \eqref{tucker_rank2} so that the next compression ratio of $(\alpha+1)$ is achieved:
\begin{equation}\label{tucker_rank2}
R_{min}=\frac{-\frac{C+\beta S}{\beta k^2} + \sqrt{\frac{(C+\beta S)^2}{\beta^2k^4} + \frac{4CS}{\beta (\alpha+1)}}}{2}
\end{equation}

\subsection{Layer Freezing}
According to Fig.\ref{fig:1a}, after applying LRD, each of the layers is decomposed into two or three layers with less number of parameters. 
However, since the weights of the new layers are calculated in closed form using \eqref{svd2} and \eqref{tucker2} so that the reconstruction error $||\W-\W'||^2$ is minimized, we can assume that only one of the decomposed layers needs to be fine-tuned and the rest of them could be considered as activation functions.
Therefore, we propose to freeze one of the two decomposed layers in SVD decomposition and first and last $1\times1$ layers in Tucker decomposition and only fine-tune the weights of the other layer in SVD and the core tensor in Tucker decomposition. 
This way, we could save a significant time during training as the number of trainable layer in the decomposed model is the same as the original model.
However, since the model architecture is unchanged, during the inference we will not gain any acceleration. 
Therefore, the freezing method is useful only in the cases where minimizing the training time has a high priority, for example for huge Transformer-based models with billions of parameters.

An advanced version of layer freezing would be to sequentially freeze/unfreeze the decomposed layers every other epoch. 
This way, all of the decomposed layers will have the chance to be fine-tuned while the number of trainable layers stays the same as the original model at each epoch.
In this method, the layers that have been frozen during the previous epoch will be unfrozen but the rest of the decomposed layers will be frozen.
This process is repeated at the end of each epoch. 
We call this technique Sequential Freezing to distinguish it from the regular freezing described in previous paragraph in which the freezing is applied onlyt once to some fixed layers.
The sequential freezing is explained in Algorithm \ref{seq-freezing} as a pseudo-code.
In this pseudo-code, the decomposed layers are shown as $L_r(i)$ where $i=0,1,2$ for regular convolutional layers and $i=0,1$ for $1\times1$ convolutional and fully connected layers. 

\begin{algorithm}
\caption{Sequential freezing of decomposed layers}
\label{seq-freezing}
    \begin{algorithmic} 
    \STATE \textbf{Input}: Decomposed Layers $L_r(i)$, Epoch number $e$
    \IF{$e\%2 = 0$}
        \IF {Tucker}
            \STATE Freeze $L_r(0)$ and $L_r(2)$ 
            \STATE Unfreeze $L_r(1)$ 
        \ELSIF {SVD}
            \STATE Freeze $L_r(0)$
            \STATE Unfreeze $L_r(1)$
        \ENDIF
    \ELSE
        \IF {Tucker}
            \STATE Unfreeze $L_r(0)$ and $L_r(2)$ 
            \STATE  Freeze $L_r(1)$ 
        \ELSIF {SVD} 
            \STATE Unfreeze $L_r(0)$
            \STATE Freeze $L_r(1)$
        \ENDIF
    \ENDIF
    \end{algorithmic}
\end{algorithm}

\section{Experiments}
In order to show how these techniques work on different types of models and AI devices, we perform experiments for convolutional neural networks on V100 NVIDIA GPUs and transformer-based models on Huawei's Ascend-910 NPUs.
We use ResNet architectures including ResNet-50, -101 and -152 for experiments performed on V100 NVIDIA GPUs \cite{he2016identity}. 
On Huawei's Ascend-910 NPUs, we use ViT model, which uses transformer modules for image classification \cite{dosovitskiy2020image}.  
In both ResNet and ViT cases, we first load the ImageNet pretrained weights into the models and then apply LRD with different proposed techniques. 
The decomposed models are then fine-tuned on ImageNet or CIFAR-10 datasets. 
For rank optimization using Algorithm \ref{rank-selection}, we start from the ranks calculated in vanilla LRD for the desired compression ratio of 2x using \eqref{tucker_rank}.
We then calculate $R_{min}$ using \eqref{tucker_rank2} so that the next integer compression ratio of (2+1)x = 3x is achieved.

For ResNet architectures, all of the convolutional and fully connected layers are decomposed used LRD. 
We then finetune the decomposed model for 45 epochs on ImageNet using a cosine learning rate scheduler and 30 epochs on CIFAR-10 using a fixed learning rate of 0.001.
The SGD optimizer with momentum 0.9 and weight decay of 1e-4 is also added to prevent overfitting. 
We calculate the average time per step over an epoch as a measure of throughput.
The throughput improvement during training and inference is reported in Table~\ref{table1} for different ResNet architectures. 
\begin{table}[h]
    \centering
    \caption{Training and inference speed-up and accuracy of ResNet-50, -101 and -152 before and after applying LRD with 2x compression and different acceleration methods.}
{\footnotesize
    \begin{tabular}{l| c c | c c}
{\bf Method} &\makecell{\bf Train \\ \bf Speed (fps)} &\makecell{\bf Train \\ \bf $\Delta$ Speed (\%)} &\makecell{\bf Infer \\ \bf Speed (fps)} &\makecell{\bf Infer \\ \bf $\Delta$ Speed}\\
\hline
\textbf{ResNet-50} &346 &0 &1232 &0\\
LRD &367 &+06.07 &1316 &+06.82\\
Rank Opt. &432 &+24.86 &1560 &+26.62\\ 
Freezing &431 &+24.57 &1316 &+06.82\\
Combined &505 &+45.95 &1560 &+26.62\\
\hline
\textbf{ResNet-101}  &207 &0 &713 &0\\
LRD &227 &+09.66 &788 &+10.52\\
Rank Opt. &282 &+36.23 &982 &+37.73\\ 
Freezing &269 &+29.95 &788 &+10.52\\
Combined &332 &+60.39 &982 &+37.73\\
\hline
\textbf{ResNet-152}  &145 &0 &510 &0\\
LRD &162 &+11.73 &577 &+13.14 \\
Rank Opt. &201 &+38.62 &694 &+36.08\\ 
Freezing &191 &+31.72 &577 &+13.14\\
Combined &232 &+60.00 &694 &+36.08\\
\end{tabular}
    }
    \label{table1}
\end{table}
 
As seen in Table~\ref{table1}, the vanilla LRD with 2x compression can improve the throughput of ResNet-50, -101 and -152 by only 6\%, 10\% and 13\%, respectively, although the number of parameters shrinks by 2 times. 
However, by applying rank optimization algorithm on top of LRD, the training (inference) speed improves by \%25 (27\%), 36\% (37\%) and 38\% (36\%) for ResNet-50, -101 and -152 models, respectively. 

On the other hand, the freezing technique can improve the training throughput by 25\%, 30\% and 32\% for for ResNet-50, -101 and -152 models, respectively.
The improvement is larger for deeper models such as ResNet-152. 
However, as seen in this table, the inference throughput stays the same as the vanilla LRD because freezing helps only regarding the backpropagation time during training  not inference. 
Table~\ref{table1}, when combined together, rank optimization and freezing can improve the training speed by 46\%, 60\% and 60\% for ResNet-50, -101 and -152 models, respectively.
The speed improvement is more significant for deeper networks e.g. ResNet-152. 

In order to show the computational complexity of the LRD, the decomposition time of the vanilla LRD with and without rank optimization and freezing techniques is compared in Table~\ref{table3} for ResNet-50, -101 and -152 models. 
As seen in this table, vanilla LRD takes around 232 seconds to decompose the ResNet-152 model while the rank optimization technique takes 716 seconds for calculating the optimal ranks. 
The overhead time the rank optimization technique adds to the total training time is in the order of minuets which is negligible comparing to the total training time of the deep learning models which is in the order of hours and even days. 
It is also clear that freezing does not add any overhead as it is applied just by setting the "requires grad" argument to True or False in the decomposed layers. 
\begin{table}[h]
    \centering
    \caption{Decomposition time of ResNet architectures using vanilla LRD with and without rank optimization and freezing on NVIDIA GPUs.}
{\footnotesize
    \begin{tabular}{l| c c c c}
\makecell{\bf Decomposition \\ \bf Time (sec)} &\makecell{\bf Vanilla \\ \bf LRD} &\makecell{\bf Rank \\ \bf Optimization} \makecell{\bf Freezing}\\
\hline
{ResNet-50} &30 &264 &30\\
{ResNet-101} &164 &489 &164\\
{ResNet-152} &232 &716 &232\\
\end{tabular}
    }
    \label{table3}
\end{table}

The accuracy of the aforementioned models is also reported in Table~\ref{table2} on two datasets including ImageNet and CIFAR-10 \cite{russakovsky2015imagenet, krizhevsky2009learning}. 
As seen in this table, the accuracy decreases slightly from vanilla LRD to rank optimization and freezing.
When combining the two techniques, the accuracy reaches it lowest value while providing the highest training speed. 
On ImageNet, rank optimization causes a minor accuracy decrement of 0.16\%, 0.06\% and 0.04\% for ResNet-50, -101 and -152 models, respectively comparing to that of vanilla LRD. 
Freezing also causes a minor accuracy decrement of 0.33\%, 0.10\% and 0.52\% for ResNet-50, -101 and -152 models, respectively comparing to that of vanilla LRD. 
As seen here, the accuracy could be recovered easily after applying these techniques. For the CIFAR-10 dataset we also see a similar trend in which all the accuracy numbers are in a the vacinity of the LRD.   

\begin{figure}
        \centering
        \includegraphics[width=100mm,scale=1]{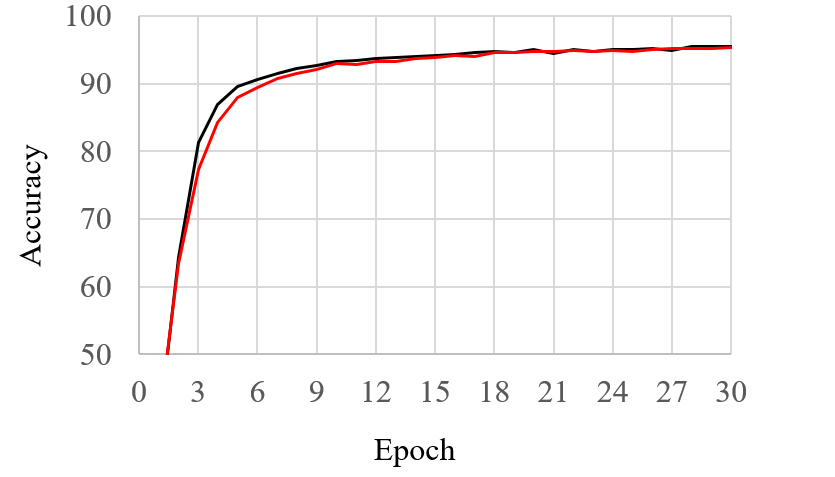}
    \caption{Fine-tuning of ResNet-50 using sequential layer freezing (black line) vs regular freezing (red line) on CIFAR-10 dataset. }
    \label{seq_freez}
\end{figure}

Fine-tuning of ResNet-50 using sequential layer freezing and regular freezing on CIFAR-10 dataset is also depicted in Fig.~\ref{seq_freez}. 
According to this figure, it can be concluded that sequential freezing leads to a faster convergence comparing to the regular freezing. 
For example, the sequential freezing reaches to accuracy of 95\% at epoch 20 while regular freezing reaches to that accuracy at epoch 26, resulting around 30\% faster convergence. 
The final accuracy of sequential freezing is also slightly higher than that of regular freezing (95.46\% vs 95.27\%) on CIFAR-10 dataset. 

\begin{table}[h]
    \centering
    \caption{Accuracy of ResNet-50, -101 and -152 on ImageNet and CIFAR-10 datasets before and after applying LRD with 2x compression and different acceleration methods on V100 NVIDIA GPUs..}
{\footnotesize
    \begin{tabular}{l|  c c c}
{\bf Method} &\makecell{\bf Accuracy \\ \bf ImageNet} &\makecell{\bf Accuracy \\ \bf CIFAR-10} &\makecell{\bf Training \\ \bf Speed-up (\%)}\\
\hline
\textbf{ResNet-50} &76.13 &96.40 &0\\
LRD &76.67 &96.01 &+06.07\\
Rank Opt. &76.51 &95.93 &+24.86\\
Freezing &76.34 &95.14 &+24.57 \\
Combined &75.96 &94.28 &+45.95\\
\hline
\textbf{ResNet-101}  &77.37 &97.60 &0\\
LRD &76.94 &97.54 &+09.66\\
Rank Opt.&76.88 &97.46 & +36.23\\
Freezing &76.84 &97.37 &+29.95\\
Combined &76.01 &96.65 &+60.39\\
\hline
\textbf{ResNet-152}  &78.31 &98.72 &0\\
LRD &77.91 &98.64 &+11.73 \\
Rank Opt.&77.87 &98.49 &+38.62\\
Freezing &77.83 &98.12 &+31.72\\
Combined &77.05 &97.29 &+60.00\\
\end{tabular}
    }
    \label{table2}
\end{table}

In order to evaluate the effect of the proposed techniques on different architectures, we applied them to the ViT model with 12 transformer modules. 
In each of the modules, there are 2 fully connected layers inside the feed forward layers that we decompose into 2 fully connected layers using SVD. 
We also decompose the fully connected layers inside the embedding layer. 
The ImageNet weights are first loaded into the original model and then it is fine-tuned on the CIFAR-10 after applying the LRD with different techniques. 
The accuracy and throughput experiments are reported in Table~\ref{table2}.  
These experiments are performed on Huawe's Ascend 910 NPUs. 
As seen in this table, we observe a similar trend we saw for ResNet architectures. 

\begin{table}[h]
    \centering
    \caption{Accuracy of ViT on CIFAR-10 datasets before and after applying LRD with 2x compression and different acceleration methods on Huawei's Ascend 910 NPUs.}
{\footnotesize
    \begin{tabular}{l|  c c c c}
{\bf Method} &\makecell{\bf Accuracy \\ \bf CIFAR-10} &\makecell{\bf Training \\ \bf Speed (fps)} &\makecell{\bf Training \\ \bf Speed-up (\%)}\\
\hline
\textbf{Org} &98.22 &1232 &0\\
LRD &98.12 &1377 &+11.79\\
Rank Opt. &98.08 &1607 &+30.44\\
Freezing &97.87 &1561 &+26.73\\
Combined &97.24 &1745 &+41.67\\
\end{tabular}
    }
    \label{table2}
\end{table}

\section{Conclusion}
In this work, we proposed two techniques for accelerating LRD models including rank optimization and layer freezing.
We showed how these methods accelerate training and/or inference without compromising the accuracy. 
Specifically, layer freezing and its advanced version sequential layer freezing accelerates the training while rank optimization could be applied for accelerating both training and inference. 
We observed minimal accuracy loss for different ResNet architectures on two commonly used datasets including ImageNet and CIFAR-10. 
We also observed that by combining these two techniques, we can reach as high as 60\% speed up for some of the models. 
A future direction could be to apply rank optimization technique to the entire model at the same time not layer by layer as proposed here.  
The rank selection criteria can also be improved in order to consider other factors such as reconstruction error. 
This can be performed in a reinforcement learning pipeline.




\bibliography{acml22}

\begin{thebibliography}{29}
\providecommand{\natexlab}[1]{#1}
\providecommand{\url}[1]{\texttt{#1}}
\expandafter\ifx\csname urlstyle\endcsname\relax
  \providecommand{\doi}[1]{doi: #1}\else
  \providecommand{\doi}{doi: \begingroup \urlstyle{rm}\Url}\fi

\bibitem[Ahmed et~al.(2022)Ahmed, Hajimolahoseini, Wen, and
  Liu]{walid2022speeding}
Walid Ahmed, Habib Hajimolahoseini, Austin Wen, and Yang Liu.
\newblock Speeding up resnet architecture with layers targeted low rank
  decomposition.
\newblock In \emph{Edge Intelligence Workshop}, 2022.

\bibitem[Bie et~al.(2019)Bie, Venkitesh, Monteiro, Haidar, Rezagholizadeh,
  et~al.]{bie2019simplified}
Alex Bie, Bharat Venkitesh, Joao Monteiro, Md~Haidar, Mehdi Rezagholizadeh,
  et~al.
\newblock A simplified fully quantized transformer for end-to-end speech
  recognition.
\newblock \emph{arXiv preprint arXiv:1911.03604}, 2019.

\bibitem[Chen et~al.(2019)Chen, Fan, Xu, Yan, Kalantidis, Rohrbach, Yan, and
  Feng]{chen2019drop}
Yunpeng Chen, Haoqi Fan, Bing Xu, Zhicheng Yan, Yannis Kalantidis, Marcus
  Rohrbach, Shuicheng Yan, and Jiashi Feng.
\newblock Drop an octave: Reducing spatial redundancy in convolutional neural
  networks with octave convolution.
\newblock In \emph{Proceedings of the IEEE/CVF International Conference on
  Computer Vision}, pages 3435--3444, 2019.

\bibitem[Cheng et~al.(2017{\natexlab{a}})Cheng, Wu, Leng, Wang, and
  Hu]{cheng2017quantized}
Jian Cheng, Jiaxiang Wu, Cong Leng, Yuhang Wang, and Qinghao Hu.
\newblock Quantized cnn: A unified approach to accelerate and compress
  convolutional networks.
\newblock \emph{IEEE transactions on neural networks and learning systems},
  29\penalty0 (10):\penalty0 4730--4743, 2017{\natexlab{a}}.

\bibitem[Cheng et~al.(2018)Cheng, Wang, Li, Hu, and Lu]{cheng2018recent}
Jian Cheng, Pei-song Wang, Gang Li, Qing-hao Hu, and Han-qing Lu.
\newblock Recent advances in efficient computation of deep convolutional neural
  networks.
\newblock \emph{Frontiers of Information Technology \& Electronic Engineering},
  19\penalty0 (1):\penalty0 64--77, 2018.

\bibitem[Cheng et~al.(2017{\natexlab{b}})Cheng, Wang, Zhou, and
  Zhang]{cheng2017survey}
Yu~Cheng, Duo Wang, Pan Zhou, and Tao Zhang.
\newblock A survey of model compression and acceleration for deep neural
  networks.
\newblock \emph{arXiv preprint arXiv:1710.09282}, 2017{\natexlab{b}}.

\bibitem[De~Lathauwer et~al.(2000{\natexlab{a}})De~Lathauwer, De~Moor, and
  Vandewalle]{de2000best}
Lieven De~Lathauwer, Bart De~Moor, and Joos Vandewalle.
\newblock On the best rank-1 and rank-(r 1, r 2,..., rn) approximation of
  higher-order tensors.
\newblock \emph{SIAM journal on Matrix Analysis and Applications}, 21\penalty0
  (4):\penalty0 1324--1342, 2000{\natexlab{a}}.

\bibitem[De~Lathauwer et~al.(2000{\natexlab{b}})De~Lathauwer, De~Moor, and
  Vandewalle]{de2000multilinear}
Lieven De~Lathauwer, Bart De~Moor, and Joos Vandewalle.
\newblock A multilinear singular value decomposition.
\newblock \emph{SIAM journal on Matrix Analysis and Applications}, 21\penalty0
  (4):\penalty0 1253--1278, 2000{\natexlab{b}}.

\bibitem[Dean et~al.(2012)Dean, Corrado, Monga, Chen, Devin, Le, Mao, Ranzato,
  Senior, Tucker, et~al.]{dean2012large}
Jeffrey Dean, Greg~S Corrado, Rajat Monga, Kai Chen, Matthieu Devin, Quoc~V Le,
  Mark~Z Mao, Marc’Aurelio Ranzato, Andrew Senior, Paul Tucker, et~al.
\newblock Large scale distributed deep networks.
\newblock 2012.

\bibitem[Dosovitskiy et~al.(2020)Dosovitskiy, Beyer, Kolesnikov, Weissenborn,
  Zhai, Unterthiner, Dehghani, Minderer, Heigold, Gelly,
  et~al.]{dosovitskiy2020image}
Alexey Dosovitskiy, Lucas Beyer, Alexander Kolesnikov, Dirk Weissenborn,
  Xiaohua Zhai, Thomas Unterthiner, Mostafa Dehghani, Matthias Minderer, Georg
  Heigold, Sylvain Gelly, et~al.
\newblock An image is worth 16x16 words: Transformers for image recognition at
  scale.
\newblock \emph{arXiv preprint arXiv:2010.11929}, 2020.

\bibitem[Gusak et~al.(2019)Gusak, Kholiavchenko, Ponomarev, Markeeva,
  Oseledets, and Cichocki]{gusak2019musco}
Julia Gusak, Maksym Kholiavchenko, Evgeny Ponomarev, Larisa Markeeva, Ivan
  Oseledets, and Andrzej Cichocki.
\newblock Musco: Multi-stage compression of neural networks.
\newblock \emph{arXiv preprint arXiv:1903.09973}, 2019.

\bibitem[Hajimolahoseini et~al.()Hajimolahoseini, Rezagholizadeh, Partovinia,
  Tahaei, Awad, and Liu]{hajimolahoseinicompressing}
Habib Hajimolahoseini, Mehdi Rezagholizadeh, Vahid Partovinia, Marzieh Tahaei,
  Omar~Mohamed Awad, and Yang Liu.
\newblock Compressing pre-trained language models using progressive low rank
  decomposition.

\bibitem[Hajimolahoseini et~al.(2018)Hajimolahoseini, Hashemi, and
  Redfearn]{hajimolahoseini2018ecg}
Habib Hajimolahoseini, Javad Hashemi, and Damian Redfearn.
\newblock Ecg delineation for qt interval analysis using an unsupervised
  learning method.
\newblock In \emph{2018 IEEE International Conference on Acoustics, Speech and
  Signal Processing (ICASSP)}, pages 2541--2545. IEEE, 2018.

\bibitem[Hajimolahoseini et~al.(2019)Hajimolahoseini, Redfearn, and
  Krahn]{hajimolahoseini2019deep}
Habib Hajimolahoseini, Damian Redfearn, and Andrew Krahn.
\newblock A deep learning approach for diagnosing long qt syndrome without
  measuring qt interval.
\newblock In \emph{Advances in Artificial Intelligence: 32nd Canadian
  Conference on Artificial Intelligence, Canadian AI 2019, Kingston, ON,
  Canada, May 28--31, 2019, Proceedings 32}, pages 440--445. Springer, 2019.

\bibitem[Hajimolahoseini et~al.(2022{\natexlab{a}})Hajimolahoseini, Ahmed,
  Rezagholizadeh, Partovinia, and Liu]{hajimolahoseinistrategies}
Habib Hajimolahoseini, Walid Ahmed, Mehdi Rezagholizadeh, Vahid Partovinia, and
  Yang Liu.
\newblock Strategies for applying low rank decomposition to transformer-based
  models.
\newblock In \emph{36th Conference on Neural Information Processing Systems
  (NeurIPS2022)}, 2022{\natexlab{a}}.

\bibitem[Hajimolahoseini et~al.(2022{\natexlab{b}})Hajimolahoseini, Redfearn,
  and Hashemi]{hajimolahoseini2022long}
Habib Hajimolahoseini, Damian~P Redfearn, and Javad Hashemi.
\newblock Long qt syndrome diagnosis and classification, May~31
  2022{\natexlab{b}}.
\newblock US Patent 11,344,246.

\bibitem[Hajimolahoseini et~al.(2023)Hajimolahoseini, Kumar, and
  Gordon]{hajimolahoseini2023methods}
Habib Hajimolahoseini, Kaushal Kumar, and DENG Gordon.
\newblock Methods, systems, and media for computer vision using 2d convolution
  of 4d video data tensors, April~20 2023.
\newblock US Patent App. 17/502,588.

\bibitem[He et~al.(2016)He, Zhang, Ren, and Sun]{he2016identity}
Kaiming He, Xiangyu Zhang, Shaoqing Ren, and Jian Sun.
\newblock Identity mappings in deep residual networks.
\newblock In \emph{European conference on computer vision}, pages 630--645.
  Springer, 2016.

\bibitem[Hinton et~al.(2015)Hinton, Vinyals, and Dean]{hinton2015distilling}
Geoffrey Hinton, Oriol Vinyals, and Jeff Dean.
\newblock Distilling the knowledge in a neural network.
\newblock \emph{arXiv preprint arXiv:1503.02531}, 2015.

\bibitem[Krizhevsky et~al.(2009)Krizhevsky, Hinton,
  et~al.]{krizhevsky2009learning}
Alex Krizhevsky, Geoffrey Hinton, et~al.
\newblock Learning multiple layers of features from tiny images.
\newblock 2009.

\bibitem[Li et~al.(2021)Li, Mesbahi, Kobyzev, Rashid, Mahmud, Anchuri,
  Hajimolahoseini, Liu, and Rezagholizadeh]{li2021short}
Tianda Li, Yassir~El Mesbahi, Ivan Kobyzev, Ahmad Rashid, Atif Mahmud, Nithin
  Anchuri, Habib Hajimolahoseini, Yang Liu, and Mehdi Rezagholizadeh.
\newblock A short study on compressing decoder-based language models.
\newblock \emph{arXiv preprint arXiv:2110.08460}, 2021.

\bibitem[Luo et~al.(2018)Luo, Zhang, Zhou, Xie, Wu, and Lin]{luo2018thinet}
Jian-Hao Luo, Hao Zhang, Hong-Yu Zhou, Chen-Wei Xie, Jianxin Wu, and Weiyao
  Lin.
\newblock Thinet: pruning cnn filters for a thinner net.
\newblock \emph{IEEE transactions on pattern analysis and machine
  intelligence}, 41\penalty0 (10):\penalty0 2525--2538, 2018.

\bibitem[Mao et~al.(2017)Mao, Han, Pool, Li, Liu, Wang, and
  Dally]{mao2017exploring}
Huizi Mao, Song Han, Jeff Pool, Wenshuo Li, Xingyu Liu, Yu~Wang, and William~J
  Dally.
\newblock Exploring the regularity of sparse structure in convolutional neural
  networks.
\newblock \emph{arXiv preprint arXiv:1705.08922}, 2017.

\bibitem[Prato et~al.(2019)Prato, Charlaix, and Rezagholizadeh]{prato2019fully}
Gabriele Prato, Ella Charlaix, and Mehdi Rezagholizadeh.
\newblock Fully quantized transformer for machine translation.
\newblock \emph{arXiv preprint arXiv:1910.10485}, 2019.

\bibitem[Rashid et~al.(2021)Rashid, Lioutas, and
  Rezagholizadeh]{rashid2021mate}
Ahmad Rashid, Vasileios Lioutas, and Mehdi Rezagholizadeh.
\newblock Mate-kd: Masked adversarial text, a companion to knowledge
  distillation.
\newblock \emph{arXiv preprint arXiv:2105.05912}, 2021.

\bibitem[Russakovsky et~al.(2015)Russakovsky, Deng, Su, Krause, Satheesh, Ma,
  Huang, Karpathy, Khosla, Bernstein, et~al.]{russakovsky2015imagenet}
Olga Russakovsky, Jia Deng, Hao Su, Jonathan Krause, Sanjeev Satheesh, Sean Ma,
  Zhiheng Huang, Andrej Karpathy, Aditya Khosla, Michael Bernstein, et~al.
\newblock Imagenet large scale visual recognition challenge.
\newblock \emph{International journal of computer vision}, 115\penalty0
  (3):\penalty0 211--252, 2015.

\bibitem[Van~Loan(1987)]{van1987matrix}
Charles Van~Loan.
\newblock Matrix computations and signal processing.
\newblock Technical report, Cornell University, 1987.

\bibitem[Zhang et~al.(2018)Zhang, Ye, Zhang, Tang, Wen, Fardad, and
  Wang]{zhang2018systematic}
Tianyun Zhang, Shaokai Ye, Kaiqi Zhang, Jian Tang, Wujie Wen, Makan Fardad, and
  Yanzhi Wang.
\newblock A systematic dnn weight pruning framework using alternating direction
  method of multipliers.
\newblock In \emph{Proceedings of the European Conference on Computer Vision
  (ECCV)}, pages 184--199, 2018.

\bibitem[Zhuang et~al.(2018)Zhuang, Tan, Zhuang, Liu, Guo, Wu, Huang, and
  Zhu]{zhuang2018discrimination}
Zhuangwei Zhuang, Mingkui Tan, Bohan Zhuang, Jing Liu, Yong Guo, Qingyao Wu,
  Junzhou Huang, and Jinhui Zhu.
\newblock Discrimination-aware channel pruning for deep neural networks.
\newblock \emph{Advances in neural information processing systems}, 31, 2018.

\end{thebibliography}






\end{document}